\documentclass[pdflatex,sn-mathphys-num]{sn-jnl}
\usepackage{graphicx}%
\usepackage{bm} 
\usepackage{multirow}%
\usepackage{amsmath,amssymb,amsfonts}%
\usepackage{amsthm}%
\usepackage{mathrsfs}%
\usepackage[title]{appendix}%
\usepackage{xcolor}%
\usepackage{textcomp}%
\usepackage{manyfoot}%
\usepackage{booktabs}%
\usepackage{algorithm}%
\usepackage{algorithmicx}%
\usepackage[noend]{algpseudocode}
\usepackage{listings}%
\usepackage{colortbl}
\usepackage{fontawesome}
\usepackage{mathrsfs}
\usepackage{subcaption} 
\usepackage{wasysym} 
\usepackage{threeparttable}
\usepackage{pifont}
\usepackage{multirow} 
\usepackage{lmodern}
\usepackage[12pt]{extsizes}

\newtheorem{theorem}{Theorem}
\makeatletter
\def\algbackskip{\hskip-\ALG@thistlm}
\makeatother
\newtheorem{lemma}[theorem]{Lemma}
\usepackage{enumitem}
\setlist[itemize,1]{label=$\times$}
\setlist[itemize,2]{label=$\checkmark$}
\setlist[itemize,3]{label=$\diamond$}
\setlist[itemize,4]{label=$\bullet$}

\usepackage{geometry}
\begin{document}


\title[ICARE: Cross-Domain Text Classification with Incremental Class-Aware Representation and Distillation Learning]{ICARE: Cross-Domain Text Classification with Incremental Class-Aware Representation and Distillation Learning}


\author[1]{\fnm{Son} \sur{Duong}}\email{son.dc232248M@sis.hust.edu.vn}
\equalcont{These authors contributed equally to this work.}

\author[2]{\fnm{Truong X.} \sur{Tran}}\email{truong.tran@psu.edu}
\equalcont{These authors contributed equally to this work.}

\author*[1]{\fnm{Hai-Anh} \sur{Tran}}\email{anhth@soict.hust.edu.vn}
\equalcont{These authors contributed equally to this work.}

\affil*[1]{\orgdiv{School of Information and Communications Technology}, \orgname{Hanoi University of Science and Technology}, \orgaddress{\street{Dai Co Viet}, \city{Hanoi}, \postcode{10000}, \country{Vietnam}}}

\affil[2]{\orgdiv{School of Science, Engineering and Technology}, \orgname{Penn State Harrisburg, The Pennsylvania State University}, \orgaddress{\state{PA}, \country{USA}}}


\abstract{The rise of powerful, fine-tuned large language models (LLMs) necessitates robust natural language inference (NLI) systems. These NLI systems ideally train across various tasks, encompassing diverse concepts like sentiment analysis, news classification, and coreference resolution. However, this continual, multi-tasking approach can lead to catastrophic forgetting, where previously learned tasks are lost, especially in domains lacking specific labels. Rehearsing past examples with labels can further exacerbate this issue by causing overfitting, hindering the system's ability to classify new categories. To tackle these challenges, this research proposes ICARE, a novel framework with two key components: an instance selection module and a knowledge distillation method. The former is based on representation learning, which meticulously selects training examples to present to the NLI system. The latter, knowledge distillation with feature and logit-level transfer, facilitates knowledge transfer between tasks. This leverages pruning techniques and optimized objective functions to transfer both essential feature-level and decision-making (logit-level) knowledge, mitigating catastrophic forgetting while adapting to new patterns. Extensive evaluations on benchmark text classification datasets have demonstrated ICARE's effectiveness. The framework achieves a significant improvement in target accuracy and reduction in the ``forgetting metric''. These results solidify ICARE's ability to preserve and leverage learned knowledge across diverse tasks, ultimately enhancing the adaptability and performance of NLI systems.

The source code is available at \url{https://github.com/CongSon01/ICARE_NLP.git}.}




\maketitle
\section{Introduction}
Natural language processing (NLP) models, particularly those powered by large language models (LLMs), have achieved remarkable progress in recent years \cite{cheng2024exploring, su2024large, xi2023rise}. Text classification tasks are evolving beyond basic functionalities like spam detection or sentiment analysis within a single domain but now demand the ability to learn continuously in an ever-evolving world \cite{biesialska2020continual}. The ability to learn and adapt across different domains is crucial in today's dynamic world.  For example, a news classifier needs to handle emerging topics like COVID-19 or cryptocurrency news that weren't present in the initial training data. This challenge has led to the rise of domain adaptation (DA) techniques \cite{ramponi2020neural, alqahtani2023comparative}. DA aims to enhance trained models from source domains to perform effectively in a target domain. Within DA, cross-domain text classification (CDTC) specifically focuses on effectively classifying text in a target domain while leveraging knowledge learned from source domains.

Catastrophic Forgetting presents a challenge in CDTC, as deep neural networks (DNNs) trained on new target domain data often forget the classification tasks learned from source domains \cite{wang2024comprehensive}. This forgetting hinders the effectiveness of CDTC, as the model needs to retain knowledge from both source and target domains for accurate classification across domains. To address this challenge of catastrophic forgetting in CDTC, incremental learning (IL) has emerged as a promising approach \cite{wang2024comprehensive, bhatt2024preventing}. IL enables models to achieve a positive transfer of knowledge from source domains to the target domain while preserving the memory of previously learned classifications. Existing incremental learning approaches for NLP tasks can be broadly classified into two categories: replay-based methods \cite{sun2019lamol, de2019episodic} and regularization-based methods \cite{yang2019simple, huang2021continual}. Replay-based methods store and reuse old examples during new task training to maintain past knowledge. Regularization-based methods, on the other hand, impose constraints on model parameters to prevent excessive changes that could lead to forgetting prior tasks. However, both approaches can encounter limitations related to \textit{training resource constraints}. Replay methods can incur high memory costs due to storing old samples, while regularization methods can lead to extended training times.

CDTC models encounter additional challenges stemming from inherent differences in text expression between source and target domains.
This can lead to overfitting, particularly on source domain instances with labels that are specific to that domain and not generalizable to the target domain. To address this issue, several studies have proposed methods to improve model generalization, such as domain adversarial training \cite{pan2022improved, wu2021conditional}, task-specific knowledge-based approaches \cite{zhou2020sentix}, and class-aware methods \cite{yang2023continual}. These methods leverage the shared characteristics between the source and target domains to enhance the model's ability to learn transferable knowledge. However, CDTC becomes even more intricate when dealing with agnostic target domains. In these scenarios, the task in the target domain is significantly different from the tasks in the source domains, and these approaches might suffer from \textit{gradient vanishing or exploding issues}. For instance, a model trained for sentiment classification on source domains (e.g., movie reviews) might struggle when applied to a target domain for the news classification task. Traditional methods that rely on shared characteristics between domains become less effective. This is because the significant difference in tasks makes it difficult for the model to find common ground for learning. the model might prioritize learning completely new patterns for the target task, neglecting to maintain the knowledge acquired from source domains. This can lead to the model forgetting previously learned patterns entirely.

Despite the success of domain adaptation (DA) and CDTC techniques, adapting language models to continually learn and retain knowledge across different domains remains challenging. Current CDTC methods often fail to balance the model's ability to generalize to new domains while retaining previously learned information from source domains. This imbalance typically results in catastrophic forgetting, where the model's performance on prior domains degrades as it learns new tasks. Such forgetting is particularly problematic in scenarios where emerging domains lack similarity to the source data, as the model struggles to maintain relevant knowledge without extensive retraining. Furthermore, conventional methods fall short in scenarios with high memory or computational constraints, limiting their practical deployment. Given these challenges, there is a strong need for a method that ensures robust knowledge retention and efficient adaptation, even when faced with diverse and evolving domain-specific data.

To address the aforementioned challenges, we propose a novel method called \textbf{I}ncremental \textbf{C}lass-\textbf{A}ware \textbf{Re}presentation and Distillation Learning (ICARE) for cross-domain text classification. ICARE employs a two-pronged strategy that tackles both inefficient source data utilization and gradient vanishing/exploding issues. First, ICARE processes text into subword tokens and extracts semantically feature-rich vectors through a feature extractor phase utilizing backbones (e.g., RoBERTa\footnote{\url{https://huggingface.co/FacebookAI/roberta-base}}, XLNet\footnote{\url{https://huggingface.co/xlnet/xlnet-base-cased}}, BERT\footnote{\url{https://huggingface.co/google-bert/bert-base-uncased}}, and DistillBERT\footnote{\url{https://huggingface.co/distilbert/distilbert-base-uncased}}). Upon obtaining these feature vectors, a small proportion of class-aware representation examples (e.g., 5\% samples selected via a herding strategy \cite{bovenzi2023benchmarking}) are stored and rehearsed to mitigate catastrophic forgetting and inflating memory during the training. This targeted selection helps retain crucial knowledge from source domains (mitigating catastrophic forgetting) while keeping memory constraints in check. Additionally, ICARE leverages knowledge distillation techniques \cite{xu2023distilling} (both feature-level and logit-level) to transfer knowledge from the feature extractor to a lightweight classifier. This knowledge transfer helps the classifier learn effectively in the agnostic target domain. Furthermore, ICARE integrates pruning techniques and a carefully designed objective function within the distillation process. This combined approach helps prevent overfitting by restraining the growth of unnecessary parameters in the classifier, especially when dealing with the challenges of agnostic target domains. Experiments on four publicly available datasets demonstrate that ICARE outperforms existing benchmarks.

To sum up, the contributions of this work are as follows:
\begin{itemize}[label={$\bullet$}]
\item \textbf{Conceptualizing CDTC as Incremental Learning}: The work presents a novel formulation of CDTC as an incremental learning task with constraints and objectives. This reframing allows us to leverage existing techniques from incremental learning to improve performance in CDTC.

\item \textbf{The ICARE Framework}: ICARE addresses CDTC by leveraging class-aware experience representation and knowledge distillation learning. This two-pronged approach ensures both stability (retaining knowledge from source domains) and plasticity (adapting to the target domain) while efficiently managing memory constraints by curbing redundant storage of source data.
\item \textbf{Empirical Validations and Benchmarks Results}: The work presents the set of comprehensive experiments and benchmark results on four publicly available datasets commonly used in CDTC research. These datasets encompass a variety of domains to assess ICARE's generalizability.  
\end{itemize}

The remainder of this paper is organized as follows: Section \ref{section:related_work} reviews a brief discussion of related research on CDTC and incremental learning. Section \ref{sec:methodology} provides the formal definition, terminology and detailed components of the proposed ICARE framework. Section \ref{sec:experiments} presents the experimental setup considered, with corresponding results discussed in Section \ref{sec:methodology}. Section \ref{sec:conclusions} concludes the research with suggestions for future research.

\section{Related Works}
\label{section:related_work}

This section provides a concise review of relevant research in CDTC and explores techniques within incremental learning, such as representation learning and knowledge distillation.

\begin{table*}[hb]
\centering
\caption{Related work for cross-domain text classification}
\label{tab:RW_QoSRouting}
\scalebox{0.55}{
\begin{threeparttable}
\begin{tabular}{@{}ccccccc@{}}
\toprule
\textbf{Year} & \textbf{Works} & \textbf{Techniques} & \textbf{IL methods} & \textbf{Datasets} & \textbf{Encoder} & \textbf{Joint Training} \\ \midrule
2020 & SentiX (Jie Zhou et al.) & \Circle\ & Model (Fine-tune) & Yelp 2020 + Sentiment & \textit{self} & WP + EP + RP \\
2021 & IDBR (Huang et al.) & \CIRCLE\ & Data + Alg (Regularization) & \begin{tabular}[c]{@{}c@{}}AGNews + Yelp + Amazon\\  + DBPedia + Yahoo\end{tabular} & BERT & TC + DC + SP \\
2022 & AdSPT (Hui Wu et al.) & \LEFTcircle\ & Model (Prompt-tuning) & Amazon reviews & \textit{self} & TC + DC \\
2022 & AMR (Li et al.) & \Circle\ & Data & Amazon review & BERT & TC + Reg \\
2024 & RCA (Hu et al.) & \LEFTcircle\ & Alg (Regularization) & FDU-MTL+ Amazon review & CNN & TC + DC + VAT \\
2024 & TACIT (Song et al.) & \CIRCLE\ & Data + Model (Distillation) & Amazon reviews & RoBERTa & TC + VAE + KD \\ \midrule
\multicolumn{2}{c}{This paper} & \CIRCLE\ & Data + Model + Alg & Yelp + DBPedia + AGNews + Yahoo & Roberta & TC + KD \\ \bottomrule
\end{tabular}
\begin{tablenotes}
\item 
\textbf{\textit{Abbreviations:}} 
\textbf{Techniques:} 
\Circle\ Task-specific knowledge-based approach,
\LEFTcircle\ Domain adversarial training, 
\CIRCLE\ Class-aware method;
\textbf{IL methods:} 
(Data) Data-Centric methods,
(Model) Model-Centric methods,
(Alg) Algorithm-Centric methods;
\textbf{Encoder:}
(Self) Their private encoder,
(BERT) Bidirectional Encoder Representations from Transformers, 
(RoBERTa) Robustly Optimized BERT Pretraining Approach, 
(CNN) Convolutional Neural Network;
\textbf{Joint Training:}
(WP) Word prediction,
(EP) Emoticon Prediction,
(RP) Rating Prediction,
(DC) Domain Classification,
(SP) Sentence Classification,
(KD) Knowledge Distillation,
(VAT) Virtual adversarial training,
(VAE) Variational Auto Encoder,
(Reg) Regularization;
\end{tablenotes}
\end{threeparttable}
}
\end{table*}

\textbf{Cross-domain text classification.}
CDTC aims to transfer knowledge from source domains to a target domain, enhancing classification accuracy by leveraging the comprehensive knowledge learned. Earlier methodologies typically trained each domain (or topic) on separate independent models and aggregated their outputs for final predictions \cite{liu2015representation, wu2015collaborative}. The most recent CDTC methods \cite{li2022cross, zhou2020sentix, wu2022adversarial, huang2021continual} facilitate generalization by capturing key features or attributes of both source and target domains. These methodologies can be categorized into three main groups: task-specific knowledge-based approaches, domain adversarial training techniques, and class-aware methods, which demonstrate state-of-the-art (SoA) performance. 
As a summary of the reviewed works is presented in Table \ref{tab:RW_QoSRouting}. The abbreviations of the table represent an abstract overview of the original meaning of the research. In \cite{wu2022adversarial}, for instance, two loss functions are employed for sentence and domain classification, the abbreviations can be adjusted to denote "Text classification" (TC) and "Domain classification" (DC).

Task-specific knowledge-based approaches supplement task-related source knowledge within the target domain. Li et al. \cite{li2022cross} leverage Abstract Meaning Representation (AMR), which integrates text- and sentiment-driven semantic graphs for cross-domain generalization. Zhou et al. \cite{zhou2020sentix} propose an abbreviation called SENTIX, which utilizes available annotations and lexicons to re-train the language model. 

Domain adversarial training techniques aid in learning features across different domains by minimizing the difference between source and target instances. Wu et al. \cite{wu2022adversarial} propose a dual adversarial co-learning (DACL) method, which combines discriminator-based and classifier-based adversarial learning. Results demonstrate the effectiveness of this method in feature extraction via adverbial alignment. Hu et al. \cite{wu2022adversarial} introduced Regularized Conditional Alignment (RCA) as a solution to challenges encountered in CDTC (e.g., neglecting class-aware information) during adversarial alignment and scarcity of labeled data across domains. RCA aligns joint distributions of domains and classes to bolster discriminative features and integrates entropy minimization and virtual adversarial training to enhance model robustness.

Class-aware methods exploit multiple discriminative source domain labels to extract category-invariant features more effectively. Huang et al. \cite{huang2021continual} propose an Information disentanglement-based regularization (IDBR) method. IDBR separates generic and specific features from the latent feature space of the source domain and utilizes generic features to promote cross-domain generalization. Song et al. \cite{song2024tacit} introduce a feature disentanglement framework called Target Domain Agnostic Feature Disentanglement (TACIT), which employs Variational Auto-Encoders to dynamically separate robust and unrobust features. Furthermore, a feature distillation task is incorporated to promote the distinction between unrobust and robust features, compelling the former to mimic the output of the teacher.

Text-based domain adaptation refers to techniques that allow a machine learning model trained in one domain (source domain) to perform well in another domain (target domain), where the data distribution differs between the two domains. In natural language processing (NLP), domain adaptation is essential because text from different domains (e.g., legal documents vs. social media posts) often varies in vocabulary, structure, and style, making it difficult for models trained on one domain to generalize well to others. Peng et al. \cite{peng2013marginalized} propose a Marginalized Denoising Autoencoder with Graph Regularization (GmSDA) for domain adaptation, aiming to improve sentiment classification by learning domain-invariant features. The model enhances the traditional marginalized stacked denoising autoencoder (mSDA) by incorporating graph regularization to preserve the local geometric structure of the data. Wei et al. \cite{wei2016deep} introduce a Deep Nonlinear Feature Coding (DNFC) framework for unsupervised domain adaptation, which enhances the marginalized stacked denoising autoencoder (mSDA) by incorporating Maximum Mean Discrepancy (MMD) for domain divergence minimization and kernelization for nonlinear coding. DNFC addresses the limitations of mSDA by ensuring the learned features have minimal distribution discrepancy between source and target domains and by capturing nonlinear data relationships. In another work, Wei et al. \cite{wei2018feature} explore why marginalized stacked denoising autoencoders (mSDA) enhance domain adaptation tasks, focusing on adaptive regularization with Gaussian and Bernoulli dropout noise. It proposes a new variant, mSDA with multinomial dropout noise (mSDAmd), which better handles feature correlations, leading to improved adaptation performance and faster convergence. Additionally, the authors introduce the deep transferable feature coding (DTFC) framework, which minimizes domain divergence through maximum mean discrepancy to further improve unsupervised domain adaptation.

\textbf{Incremental Learning.}
The objective of IL is to achieve effective adaptation between old and new knowledge, overcoming the challenge of catastrophic forgetting while ensuring stability and plasticity. Existing IL researches \cite{zhou2023deep} could be broadly categorized into three main categories: Data-Centric, Model-Centric, and Algorithm-Centric method. The Data-Centric methods entail reviewing former classes and mitigating forgetting through experience replay, representation techniques, and regularization \cite{bang2021rainbow, de2021continual}. The Model-Centric methods directly optimize knowledge transfer among tasks or enhance representation capabilities through example representations \cite{yoon2017lifelong, li2019learn}. The Algorithm-Centric method focuses on designing algorithms to minimize interference between source and target tasks  \cite{zhou2022forward, liu2021adaptive}. Among these approaches, Data-Centric methods have found widespread application in NLP tasks, enabling continual acquisition of novel world knowledge from streams of textual sentences without forgetting \cite{radford2019language}. 

Knowledge distillation is a fundamental approach for establishing student-teacher mappings \cite{ye2021lifelong} in CIL with various formats (logits, features, and relationships). Due to their inherent flexibility, knowledge distillation-based methods have been extensively applied to a range of incremental learning tasks, including person re-identification \cite{pu2021lifelong, pu2023memorizing}, question answering \cite{wang2020efficient}, and relation extraction \cite{zhao2022consistent, huang2024dp}. However, they often suffer from large memory costs or considerable training times due to storing excessive old samples and model parameters (related to resource constraints). Prabhu et al. \cite{prabhu2020gdumb} introduce GDumb, a simple continual learning approach that outperforms many complex methods by using a greedy sampling strategy to store balanced samples and training a model from scratch only on these samples at test time, challenging the current benchmarks and assumptions in continual learning. Ye et al. \cite{ye2024self} present Self-Supervised Adversarial Variational Learning (SS-AVL), a hybrid model combining Variational Autoencoders (VAEs) and Generative Adversarial Networks (GANs) to learn interpretable and disentangled data representations without real data, using self-supervised learning and mutual information maximization to optimize the inference and generator networks separately, demonstrating improved performance in complex image generation and representation tasks. The authors also introduce the Continual Variational Autoencoder (CAA) \cite{ye2024task}, a task-free continual learning model that uses a dynamic, bio-inspired memory system to manage temporary and long-term data storage, enabling the model to generate diverse images and maintain robust representations across evolving data streams without explicit task boundaries. Zenke et al. \cite{zenke2017continual} introduce Synaptic Intelligence (SI), a continual learning approach that reduces catastrophic forgetting by tracking each synapse's importance during training, enabling the model to selectively protect critical synapses and retain previously acquired knowledge across tasks, mirroring aspects of biological neural networks.

Overall, existing CDTC research primarily focuses on scenarios where the source and target domains share some underlying similarities. These methods typically rely on identifying commonalities between source and target knowledge for effective knowledge transfer. However, this approach becomes significantly less effective when dealing with agnostic target domains, where the task in the target domain is fundamentally different from those encountered in the source domains. This limitation motivates the development of ICARE, a novel framework specifically designed for CDTC in agnostic target domains. ICARE tackles the dual challenges of catastrophic forgetting and agnostic target features by leveraging techniques from incremental learning to relearn class-aware experience representations while simultaneously utilizing previous knowledge from the source domains. In addition, while leveraging past knowledge is crucial, directly rehearsing all source domain examples can be memory-intensive. Our research aims to address this by strategically enhancing representation methods by rehearsing only a limited number of carefully selected examples during training.


\begin{figure}
    \centering
    \includegraphics[width=0.8\linewidth]{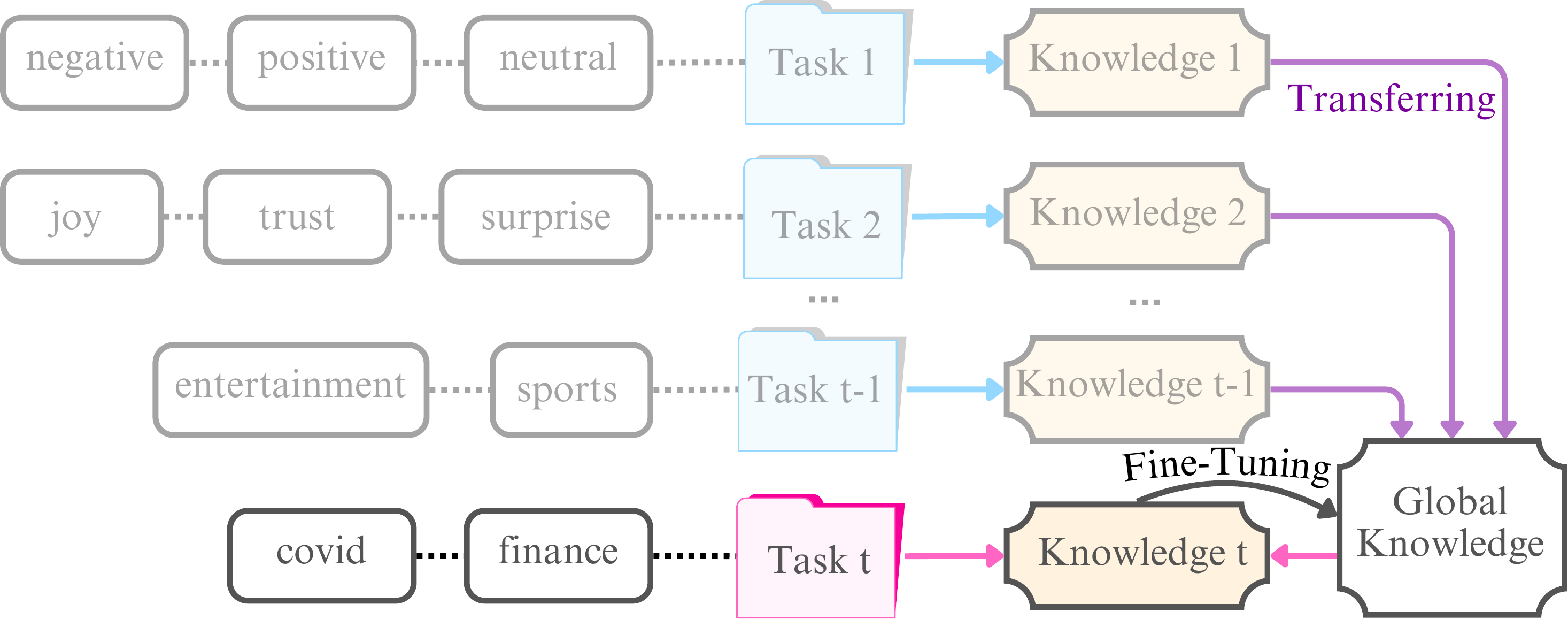}
    \caption{An example of Cross-Domain Text Classification (CDTC) problem}
    \label{fig:crossdomain}
\end{figure}

\begin{figure}[!t]
    \centering
    \includegraphics[width=1\linewidth]{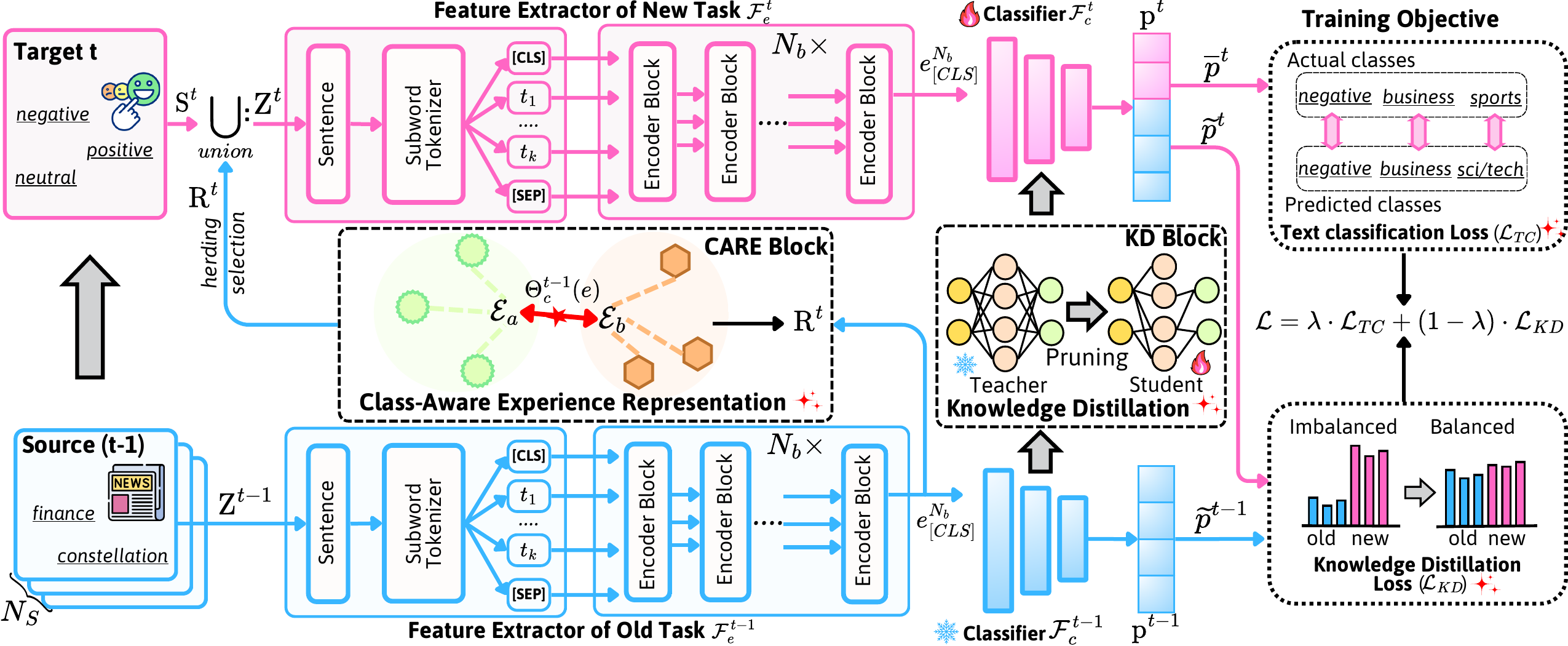}
    \caption{\textbf{ICARE framework}. It mainly contains a \textit{Class-Aware Experience Representation} block for selecting exemplars, along with \textit{Knowledge Distillation} block to transfer knowledge from the previous model $\mathcal{F}^{t-1}_c$ to the current model $\mathcal{F}^{t}_c$, and \textit{Training Objective} phase designed to balance the stability-plasticity trade-off.
    \includegraphics[height=0.2cm]{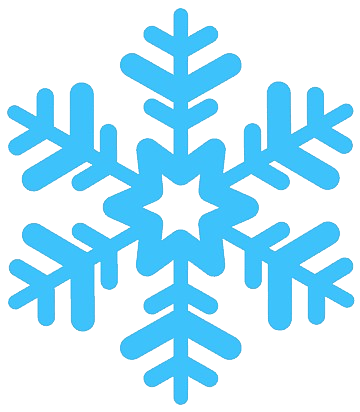} indicates that previous model parameters remain unchanged during training. \includegraphics[height=0.3cm]{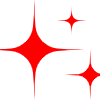} present the our key components analysing.}
    \label{fig:Methodology}
\end{figure}

\section{Methodology}
\label{sec:methodology}
This section delves into the inner workings of ICARE, the proposed framework for addressing the dual challenges of catastrophic forgetting and target-agnostic features in CDTC. The section will establish the problem formulation and lay the groundwork with key symbolic definitions. Following this foundation, a detailed exploration of ICARE's components is presented. 

\subsection{Conceptualizing CDTC as Incremental Learning and ICARE solutions}
\label{sec:Problem_Formulation}
Building upon CDTC methods \cite{wu2022adversarial, huang2021continual, song2024tacit}, a sequence of incremental learning tasks is formulated within the unsupervised domain adaptation setting, facilitating the transfer of acquired knowledge from the source domains to the target domain.

\begin{table}[hb]
    \centering
    \caption{Notation Description}
    \begin{tabular}{@{}ll@{}}
    \toprule
    \textbf{Notation} & \textbf{Description} \\ \midrule
    task $t$ & The current learning task $t$ \\
    $S^t$ & Set of new data sample (sentences) at task $t$ \\
    $R^t$ & Set of exemplar memory samples to support the training of model $\mathcal{F}^t$ \\
    $Z^t = S^t \cup  R^t$ & A combination set of $S^t$ and $R^t$ for training the $\mathcal{F}^t$ model\\
    $\textbf{x}^t_j = [w_1^t, ..., w_n^t]$ & A sentence with words $w_1^t$ to $w_n^t$ \\
    ${y}^t_j$ & The corresponding one-hot encoded class label of sentence $\textbf{x}^t_j$ \\
    $C^t$ & The class label space of task $t$. Hence, ${y}^t_j \in C^t$ \\
    $\mathcal{F}_e^t$ & The feature extractor model being trained at task $t$ \\
    $\mathcal{F}_c^t$ & The classification model being trained at task $t$ \\
    \bottomrule
    \end{tabular}
    \label{tab:my_label}
\end{table}

\textbf{Problem Definition:} Consider a CDTC problem with $T$ different tasks, and the domain (or dataset) of the $t^{th}$ task is denoted as $S^t$. A sentence-label pair $(x^t,y^t) \in S^t$ consists of $\textbf{x}^t_j = [w_1^t, ..., w_n^t]$ to represent a sentence with words $w_1^t$ to $w_n^t$  and $\textbf{y}^t_j$ to represent the corresponding one-hot encoded label belonging to the class label space $C^t$. 
The crucial aspect here is that the class label spaces of different tasks are entirely distinct. There is no overlap label between any two text classification tasks. This signifies that when considering the current task $t$ as the target, all labels in $C^t$ are new and completely different from the combined set $\bigcup_{i=1}^{t-1}C^i$ of classes learned from previous sources of the $1^{st}$ task to the $(t-1)^{th}$ task. This characteristic is also known as the agnostic target domain. 

For instance, consider the example CDTC problem layout in Figure \ref{fig:crossdomain}, where each text classification task has its unique set of class labels. The model might learn to classify a social tweet as joy, trust, or surprise, then in the next task, learn to classify news articles into different categories like entertainment or sports. The difficulty lies in training the model on these new tasks while preserving its knowledge from previous tasks. The model shouldn't ``forget" how to classify emails as spam just because it's now learning about news categories.

To manage knowledge from previously encountered classes (source domains) within the current task (where the labels are entirely different), we leverage a set of sentences called exemplar memory $R^t$. This memory set, inspired by concepts introduced in \cite{huang2024exemplar, huang2021continual}, stores a small number of representative sentences (exemplars) for each class in the previous source tasks. Importantly, the size of this memory is strictly controlled to be significantly smaller than the number of training examples in the current domain ($\left| R^t \right| \ll \left| S^t \right|$), ensuring efficient memory usage.


\textbf{Objective:} The primary objective is to train the classification model $\mathcal{F}^t)$ at the $t^{th}$ task that can accurately classify text (sentences) belonging to both the current target domain and all previous source domains. To achieve this dual goal, we aim to minimize the model's generalization error  $E(\mathcal{F}^t)$ across all tasks, ensuring it can accurately classify sentences (text) from both the new and old classes even though the target domain is agnostic (completely different labels):  

\begin{equation}
    min(E(\mathcal{F}^t)) = min(\sum_{i=1}^{T}\mathbb{E}_{(\textbf{x}^i,\textbf{y}^i) \sim \mathcal{T}^i} \mathcal{L}(\mathcal{F}(\textbf{x}^i),\textbf{y}^i))
\end{equation}

\textbf{ICARE Framework Solutions: } The framework ICARE (or Incremental Class-Aware Representation and Distillation Learning) is specifically designed to tackle the challenges of CDTC in agnostic target domains. Figure \ref{fig:Methodology} illustrates how ICARE addresses the CDTC problem when transitioning between the $(t-1)^{th}$ task (source) and $t^{th}$ task (target). As the figure depicts on the left side, the $t^{th}$ task presents an agnostic target domain, where the label space is entirely different from previous tasks. 

To navigate this challenge, ICARE leverages a two-stage model, denoted as $\mathcal{F}^t = \mathcal{F}^t_e \circ \mathcal{F}^t_c$ at the $t$-th task, which is composed of a \textit{Feature Extractor} $\mathcal{F}^t_e$ and a \textit{Task Classifier} $\mathcal{F}^t_c$. The two-stage model $\mathcal{F}^t$ for the $t^{th}$ task is illustrated at the top of Figure \ref{fig:Methodology}, while the bottom stages are for the $\mathcal{F}^t$ of the $(t-1)^{th}$ task. The final step of the ICARE framework is the \textit{Training Object} phase  (depicted on the right side of the figure), which is designed to balance the learning objective between $\mathcal{F}^t$ and its predecessor $\mathcal{F}^{t-1}$. Further descriptions of these are as follows:

\begin{itemize}[label={$\bullet$}]
\item Training Data for $\mathcal{F}^t$: To address the agnostic target domain challenge, the training data $Z^t$ for $\mathcal{F}^t$ is designed to combine the new data and some exemplar data selected from the previous task. Hence, $Z^t = S^t \cup R^t$. The set $R^t$ is referred to as exemplar memory. The selection of $R^t$ is done by the supportive block called Class-Aware Experience Representation (will be explained later).  

\item Feature Extractor $\mathcal{F}^t_e$: This stage processes the input sentences in  $Z^t$.  First, it tokenizes the sentences using subword tokenizer techniques to break them down into smaller units $\textbf{x} = [t_1, ..., t_k]$.  Next, it employs $N_b$  encoder blocks (e.g., BERT and DistillBERT embeddings) to convert these tokens into vector representations.  The resulting features are captured in $\textbf{e} = [e^{N_b}_{[\text{CLS}]}, e_1^{N_b}, ..., e_C^{N_b}, e^{N_b}_{[\text{SEP}]}]$, where  $e^{N_b}_{[\text{CLS}]}$  specifically represents the overall text output representation. 

\item Task Classifier $\mathcal{F}^t_c$ with Knowledge Distillation: The task classifier $\mathcal{F}^t_c$ is designed to classify the $e^{N_b}_{[\text{CLS}]}$ text representation. $ \mathcal{F}^t_c$ is built to learn upon the previous task  classifier $\mathcal{F}^{t-1}_c$. ICARE adds more output neurons to $\mathcal{F}^t_c$ compared to $\mathcal{F}^{t-1}_c$ to accommodate class labels of all previous tasks and new task $t$. Furthermore, ICARE employs a Knowledge Distillation block to leverage knowledge from the previous model during training, enhancing the classifier's ability to learn effectively in the agnostic target domain.

\item Training Objective: This phase is a novelty step in the ICARE framework to balance the crucial trade-off between stability and plasticity. It combines two loss functions: a standard text classification loss and a knowledge distillation loss. The classification loss guides the model towards accurate target domain classification, while the knowledge distillation loss leverages the transferred knowledge from the previous model. This combined objective function steers the model towards optimal performance in agnostic target domains.

\end{itemize}

A key aspect of ICARE's effectiveness lies in its ability to leverage previously learned knowledge for effective adaptation in agnostic target domains. To achieve this, ICARE incorporates two novel components within its framework, as illustrated in the middle section of Figure \ref{fig:Methodology}:


\begin{itemize}[label={$\bullet$}]
    \item Class-Aware Experience Representation: This block strategically selects informative examples from the source domain. It achieves this by analyzing their proximity to the average centroid (representative point) within a latent space. By prioritizing these informative exemplars, ICARE ensures efficient memory usage while retaining crucial knowledge for the target domain.
    \item Knowledge Distillation: This block facilitates the transfer of knowledge from a previously trained model ($\mathcal{F}^{t-1}_c$) to the current model ($\mathcal{F}^{t}_c$). This knowledge transfer helps the current model learn effectively in the target domain, even when it significantly differs from the source domain.
\end{itemize}

The three innovative components - Class-Aware Experience Representation, Knowledge Distillation with Classifier Expansion, and Training Objective - work in tandem to address the challenges of catastrophic forgetting and target-agnostic features, enabling the ICARE framework to excel in complex agnostic target domain scenarios. The following sections will provide the intricate details of these three novelty components of the ICARE framework.



\subsection{Class-Aware Experience Representation for Exemplar Selection}
\label{sec:Class-Aware Experience Representation}

\begin{algorithm}
\caption{Class-Aware Experience Representation}
\label{alg:Class-Aware Experience Representation}
  \textbf{Input:} The set of new sentences $S_{t}$, the set of previous sentences $Z^{t-1}$ in task $t-1$, $\gamma$ (ratio of $R^t$ to $S^t$);\\
    \textbf{Require:} Feature Extractor $\mathcal{F}^{t-1}_e$\\
    \textbf{Output:} Exemplar-set $R^t$; 
  \begin{algorithmic}[1]
  \State $R^t = \emptyset$
    \State $N^t_r = \gamma \cdot |S^t|$ \Comment{Number of exemplars in $R^t$;}
    \State $N^{old}_{class} = \left| \text{Unique}(\textbf{y}^{t-1}_j) \right| \sim C^{old}$ \Comment{Number of old classes;}
    \State $N^t_c = \left\lceil N^t_r / N^{old}_{class} \right\rceil$ \Comment{Number of exemplars per class;}
    \For{c = 1 to $N^{old}_{class}$}{}
      \State$\textbf{D}_c = \{(\textbf{x}, y) \in Z^{t-1} | y = c\}$
      \State $\mu_c = \frac{1}{\left| \textbf{D}_c \right|}\sum_{\textbf{x}\in \textbf{D}_c}\mathcal{F}_e^{t-1}(\textbf{x})$ \Comment{Cluster mean;}
        \For{k $\leftarrow$ 1 to $N^t_c$}{} \Comment{Herding selection;}
            \If {k =1}
            \State $d$ = 0
            \EndIf
            \State $x_k = \underset{\left\{ \textbf{x}, y \right\} \in \textbf{D}_c}{argmin}\left\{ \mu_c - \frac{1}{k}\left[ \mathcal{F}_e^{t-1}(\textbf{x}) + d \right] \right\}$
            \State $d$ = d + $\mathcal{F}_e^{t-1}(x_k)$
            \State $R^t \gets (x_k,c)$
        \EndFor
    \EndFor
    \State \textbf{Return} $R^t$
  \end{algorithmic}
\end{algorithm}

One of the core challenges in continual learning scenarios like agnostic target domain text classification is catastrophic forgetting. This phenomenon occurs when a model trained on multiple tasks forgets previously learned information as it adapts to new tasks. To address this challenge, the proposed Class-Aware Experience Representation module of the framework ICARE is designed to address the challenge of redundant learning and weight drift that can occur when training models across domains. In many cross-domain tasks, simply re-using all examples from the source domain for rehearsal can lead to performance degradation, as highly similar sentences may result in redundant learning \cite{bovenzi2023benchmarking}. This, in turn, causes unnecessary updates to model parameters (weight drift), ultimately affecting performance on the target domain. The purpose of this module is to strategically form an exemplar memory set by selecting a limited number of diverse and informative examples (exemplars) from the source domain for rehearsal. By prioritizing exemplars that are representative and diverse for each class, this module ensures that the learning process focuses on key instances, thus improving generalization to the target domain while minimizing the risk of weight drift.

The procedure of this block is presented in Algorithm \ref{alg:Class-Aware Experience Representation}.  


Algorithm \ref{alg:Class-Aware Experience Representation}  outlines the detailed process for constructing the exemplar memory $R^t$ for the current task $t$. The algorithm leverages the previously stored sentences $Z^{t-1}$ from source domains and the feature extractor $\mathcal{F}_e^{t-1}$  from the model in the previous task $t-1$. The algorithm aims to create a small set of exemplars with the ratio $\gamma = 0.05$ to the size of $S^t$ (the set of sentences encountered in the target domain of task $t$). The algorithm employs a hearing selection, which prioritizes samples closest to the centroid (most representative of the class) while also ensuring some level of diversity within the chosen exemplars.    

The step-by-step procedure of Algorithm \ref{alg:Class-Aware Experience Representation} is as follows:
\begin{itemize}[label={$\bullet$}]
    \item \textbf{Initialize $R^t$ as an empty set (Line 1)}: The algorithm starts by initializing an empty set to store the exemplar.
    \item \textbf{Calculating Exemplar Distribution per Class (Lines 2-4)}: This step calculates a uniform number of samples  $N_c$ for each class in the source domain to include in the exemplar memory $R^t$. It considers the desired size of $R^t$ (referred to as $N^t_r$) and the number of known classes $N^{old}_{class}$ encountered in previous source domains. This ensures an even distribution of exemplars across all classes. The calculation of $N_c$ is as follows:
    \begin{equation}
        N^t_r = \gamma \cdot |S^t| 
    \end{equation}
    \begin{equation}
      N^{old}_{class} = \left| \text{Unique}(\textbf{y}^{t-1}_j) \right| \sim C^{old}
    \end{equation}
    \begin{equation}
     N^t_c = \left\lceil N^t_r / N^{old}_{class} \right\rceil
    \end{equation}
    
    \item \textbf{Selecting Informative Exemplar per Class (Lines 5-13)}: This section focuses on selecting $N_c$ exemplar samples for each class from the source domain. The algorithm iterates over each class $c$ in the source domain and processes the following steps:  
    \begin{itemize}[label={$\diamond$}]
        \item Line 5: Get all candidate samples belonging to the current class $c$. For convenience, we denote this set of candidate samples per class as $D_c$  
        \item Line 7: Calculate the centroid $\mu_c$ (cluster mean) for each class $c$ in the latent space (a high-dimensional representation learned by the feature extractor). This centroid represents the ``center of mass" for all examples within a class:
        \begin{equation}
            \mu_c \gets \frac{1}{\left| \textbf{D}_c \right|}\sum_{(\textbf{x},y)\in \textbf{D}_y}\mathcal{F}_e^{t-1}(\textbf{x})
        \end{equation}
        \item Line 8-13: Employ the herding selection to choose $N_c$ exemplar samples of the current class $c$. The selection process favors samples closest to the centroid $\mu_c$ in the latent space. While prioritizing centrality, herding selection also ensures some level of variation among the chosen exemplars. This helps prevent the model from overfitting to a narrow set of exemplar characteristics within each class. These selected exemplars for the current class $c$ are added to the $R^t$ exemplar memory.
    \end{itemize}
    \item \textbf{Returning the exemplar memory (line 14)}. After all classes in the source domain have been considered, the algorithm returns the $R^t$ exemplar memory.
\end{itemize}

After obtaining the exemplar memory set $R^t$, ICARE combines it with $S^t$ to form the training set $Z^t = R^t \cup  S^t$ for the model at the current task $t$. 


\subsection{Knowledge Distillation with Weight Pruning}
\label{sec:Knowledge Distillation}

Catastrophic forgetting occurs when a model forgets previously learned information as it adapts to new tasks. To combat this, ICARE utilizes knowledge distillation through two complementary techniques: feature-level and logit-level distillation. In this framework, the current classification model $\mathcal{F}^{t}_c$ (student model) aims to replicate the behavior of the previous model $\mathcal{F}^{t-1}_c$ (the teacher model), while simultaneously learning new, unknown classes.
Knowledge from the teacher model is distilled and selectively pruned to form the student model (\textit{feature-level distillation}), which retains essential knowledge (e.g., identifying key features for emotion classification). The student model is then expanded and trained on new datasets, with the primary objective of replicating the predictions and probability distributions of the teacher model (\textit{logit-level distillation}) \cite{pu2023memorizing}.
The detailed procedures are outlined as follows.

\textbf{Weight Pruning for Model Efficiency:} First, ICARE employs a weight pruning technique inspired by \cite{mao2023statistical}. As the number of learning tasks grows, the model size can become unwieldy. Weight pruning can help to combat this by identifying and removing weights with minimal impact on the model's performance, effectively reducing the model size (number of training parameters) for improved efficiency. The specific implementation involves a zero-one mask matrix $M$ and a weight threshold $ThS$ to prune weights in the teacher model, as shown in the following equations:
\begin{equation}
    M[i,j] =
    \begin{cases}
      0, & \text{if}\ \left| W_{ij} \right| < ThS \\
      1, & \text{otherwise}
\end{cases}
\end{equation}

The weight parameters of the current classification model $\mathcal{F}_c^t$ is initially pruned from the previous model $\mathcal{F}_c^{t-1}$ as follow:

\begin{equation}
\tilde{W}^{(\mathcal{F}_c^t)} = W^{(\mathcal{F}_c^{t-1})} \odot M^{({\mathcal{F}}_c^{t-1})}
\end{equation}


\textbf{Knowledge Distillation Loss:} Next, ICARE facilitates the knowledge transfer between the teacher and student models by incorporating the Knowledge Distillation Loss $\mathcal{L}_{KD}$. Here's a breakdown of how $\mathcal{L}_{KD}$ achieves this knowledge transfer.   

Traditionally, models are trained using the ground-truth labels, which are one-hot encoded vectors representing the true class for each data point. These labels are essentially ``hard" targets, providing a definitive answer. However, in knowledge distillation, $\mathcal{L}_{KD}$ utilizes the similarity between the student model's soft probabilities (denoted as $\widetilde{p}^{t}$) and the teacher model's soft probabilities (denoted as $\widetilde{p}^{t-1}$). These soft probabilities are typically the distribution of probabilities across all possible classes. This ``softness" allows the student model to learn not just the most likely class (as indicated by the highest probability in the distribution) but also gain insights into the teacher's confidence level for each class. $\mathcal{L}_{KD}$ essentially quantifies the difference between two probability distributions using the Kullback-Leibler divergence (KLD, denoted as $\mathcal{D}_{KL}(.||.)$) \cite{kullback1951kullback} as depicted in Equation \ref{eq:LKD}.

\begin{equation} 
\scalebox{0.87}{$
\begin{split}
    \mathcal{L}_{KD} & \triangleq H(\widetilde{p}^{t-1}, \widetilde{p}^{t})\\
    & = \left| C^{old} + C^t \right|^{-1}\sum_{i=1}^{C^{old} + C^t}\mathcal{D}_{KL}\left(\widetilde{p}^{t-1} \left( \textbf{e}^t_i; \mathcal{F}^{t-1}; \tau \right), \widetilde{p}^{t} \left( \textbf{e}^t_i; \mathcal{F}^{t}; \tau \right) \right)
\end{split}
$}
\label{eq:LKD}
\end{equation}

In this context, a lower $\mathcal{L}_{KD}$ value indicates that the student model's predictions are getting closer to those of the teacher model, signifying successful knowledge transfer. As shown in Equation \ref{eq:LKD}, $\mathcal{L}_{KD}$ uses the distillation temperature  $\tau$ to reduce the variance of the probability distribution. A higher temperature $\tau$ softens the probability distributions of both the teacher and student models, making them smoother and easier for the student to learn from. This is particularly helpful in scenarios where the teacher model might have very high confidence (probabilities close to 1) for certain classes. By softening these probabilities, the student has a better chance of understanding the underlying reasoning behind the teacher's predictions. Conversely, a lower temperature $\tau$ allows for more focused learning, where the student primarily concentrates on replicating the teacher's most confident predictions.

\subsection{Training Objective}

\label{sec:Training Objective}

This section delves into the training objective, a crucial element in ICARE that guides the model towards optimal performance. ICARE tackles the trade-off between adapting to new data (plasticity) and retaining prior knowledge (stability) through its training objective. This objective is incorporated through a combined loss function during training: Text Classification Loss $\mathcal{L}_{TC}$ and Knowledge Distillation Loss $\mathcal{L}_{KD}$.  

Knowledge Distillation Loss $\mathcal{L}_{KD}$ (Equation \ref{eq:LKD}) has been presented in the previous section to ensure the target model inherits knowledge from previous models, mitigating catastrophic forgetting and promoting stability. The Text Classification Loss $\mathcal{L}_{TC}$ is a standard loss that measures the discrepancy between the model's predictions (represented by probabilities) and the actual ground-truth labels. While the underlying mathematical details involve cross-entropy loss, the core concept of $\mathcal{L}_{TC}$ is to minimize this discrepancy, ensuring the model aligns its predictions with the true labels for the new data.

The combined loss function for training model $\mathcal{F}^t$ is depicted in Equation \ref{eq:Loss}: 
\begin{equation}
\label{eq:Loss}
    \mathcal{L} = \lambda \cdot \mathcal{L}_{TC} + (1 - \lambda) \cdot \mathcal{L}_{KD}
\end{equation}
where $0\le \lambda \le  1$ is the weighted coefficient. The $\lambda$ value allows ICARE to balance the trade-off between adapting to new data (plasticity) and retaining prior knowledge (stability). When lambda is closer to 1, the emphasis is placed on ${L}_{TC}$ to prioritize the model's ability to classify data accurately in the new target domain.  Conversely, a lambda value closer to 0 gives more weight to $\mathcal{L}_{KD}$ emphasizes knowledge transfer from the previous model.

In addition, the cruciality of incorporating the knowledge distillation loss $\mathcal{L}_{KD}$ into the loss function $ \mathcal{L}$ for the training model of the student model can be formally proved by the smoothness characteristic of soft probabilities. For a quick recall, $\mathcal{L}_{KD}$ is computed from the difference between soft prediction probabilities of the student and teacher models, while ${L}_{TC}$ is obtained from the hard prediction probabilities. We have observed that soft prediction probabilities have lower variance than hard prediction probabilities. This characteristic is formally stated in Lemma \ref{le:variance}. This smoother guidance translates to a more stable student model that can effectively adapt to new data while retaining knowledge from previous tasks.

\begin{lemma}
The variance of hard probabilities is greater than that of soft probabilities, such that  $var(\widetilde{p}) \le var(\overline{p})$.
\label{le:variance}
\end{lemma}

\begin{proof}
To simplify, the input layer consists of two values, 0 and 1 (i.e., binary classification).\\
The equivalence regarding the variance of the probability distribution $p$ is expressed as follows:
\begin{equation}
\begin{split}
        var(p)&= \text{E}((p - \text{E}(p))^2)\\
        & = \text{E}(p^2) - \text{E}(p)^2\\
        & = \text{E}(p^2) - \frac{1}{C^2}\\
\end{split}
\end{equation}
Where $C$ is the number of classes and E represents the expected value (or expectation). Given that $\frac{1}{C^2}$ is a constant, to demonstrate $var(\widetilde{p}) \le var(\overline{p})$, we can simplify the proof to:
\begin{equation}
    \text{E}(\widetilde{p}^2) \le  \text{E}(\overline{p}^2)
\end{equation}
A sigmoid function $\sigma$ with distillation temperature $\tau$: 
\begin{equation}
    \sigma(\textbf{e}, \tau) = \frac{\text{exp}(e_i/\tau)}{\sum_{i=1}^{C^{old} +C^t}\text{exp}(e_i/\tau)}
\end{equation}
Define:
\begin{equation}
    F \triangleq \text{E}(\widetilde{p}^2) = \sum_{i=1}^{C^{old} +C^t}(\sigma(\textbf{e}, \tau))^2
\end{equation}
Herein, the derivative of the sigmoid function $\delta\sigma(x)$ is expressed as follows:
\begin{equation}
\begin{split}
    \frac{\delta\sigma(x)}{\delta x}&= \sigma(x)\cdot(1 - \sigma(x))
\end{split}
\end{equation}
Thus:
\begin{equation}
    \begin{split}
\frac{\delta F}{\delta\tau}&=\sum_{i=1}^{C^{old} +C^t}\frac{\left( \delta\sigma(\textbf{e}, \tau) \right)^2}{\delta\tau}\\
&= \sum_{i=1}^{C^{old} +C^t}\frac{\left( \delta\sigma(\textbf{e}, \tau) \right)^2}{\delta\sigma(\textbf{e}, \tau)} \cdot \frac{\delta\sigma(\textbf{e}, \tau)}{\delta\textbf{e}/\tau} \cdot \frac{\delta\textbf{e}/\tau}{\delta\tau}\\
&= \sum_{i=1}^{C^{old} +C^t}2\cdot\sigma(\textbf{e}_i,\tau)\cdot \sigma(\textbf{e}_i,\tau)\cdot(1-\sigma(\textbf{e}_i,\tau))\cdot\frac{\delta\textbf{e}/\tau}{\delta \tau}\\
&= \sum_{i=1}^{C^{old} +C^t}2\cdot \sigma(\textbf{e}_i,\tau)^2\cdot(1-\sigma(\textbf{e}_i,\tau))\cdot\frac{-\textbf{e}^2}{\tau^2}\\
\end{split}
\end{equation}
Since $\sigma(\textbf{e}, \tau) \in (0,1)$ and $\textbf{e}_i \ge 0$,
\begin{equation}
    2\cdot \sigma(\textbf{e}_i,\tau)^2\cdot(1-\sigma(\textbf{e}_i,\tau))\cdot\frac{-\textbf{e}^2}{\tau^2} \le  0, \forall \textbf{e}_i, i = \overline{1, C^{old} + C^t}
\end{equation}
That means $\frac{\delta F}{\delta \tau} \le 0$, indicating that $F$  is a decreasing function. Consequently, $\sigma(\textbf{e}, \tau) \le \sigma(\textbf{e}, 1)$, implying $\text{E}(\widetilde{p}^2) \le  \text{E}(\overline{p}^2)$, which in turn leads to $var(\widetilde{p}) \le var(\overline{p})$.
\end{proof}

\section{Experiments and Evaluations}
\label{sec:experiments}
This section presents the experiments and benchmark results for evaluating ICARE performance.   

\subsection{Experimental Setups}
\label{sec:Settings}
\subsubsection{Datasets}
\label{sec:dataset}

We evaluate our methods on six benchmark text classification datasets, covering diverse domains like news (AG News \cite{zhang2015character}, 20 Newsgroup \cite{albishre2015effective}), question-and-answer (Yahoo! Answers \cite{chang2008importance}), Wikipedia article classification (DBPedia \cite{mendes2012dbpedia}), and sentiment analysis (Yelp \cite{zhang2015character}, Amazon \cite{zhang2015character}). To ensure fair impact with respect to the amount of data in each domain, we created reduced versions of most datasets by randomly sampling data points from the original training and development sets. Additionally, to ensure the specificity of the training tasks, in the 20 Newsgroup dataset, the topics of Religion and Politics are only selected and processed, while Computer Science and Entertainment topics are utilized in the AGNews dataset. The Amazon dataset is merged with the Yelp dataset classes due to their semantic similarities \cite{de2019episodic}. Tab. \ref{tab:dataset} shows a detailed overview of four benchmark text classification datasets, including the number of classes, domain types, and the quantity of data samples in Train, Dev, and Test subsets.

\begin{table*}[h!]
\caption{\textbf{Dataset Statistics}}
\label{tab:dataset}
\centering
\begin{tabular}{@{}llllll@{}}
\toprule
\textbf{Dataset} & \textbf{Classes} & \textbf{Type of domain} & \textbf{Train} & \textbf{Dev} & \textbf{Test} \\ \midrule
Yelp    & 5 & Sentiment & 5000 & 1000 & 1000 \\
Amazon   & 5 & Sentiment   & 5000 & 1000 & 1000 \\
DBPedia & 5 & Wikipedia & 5000 & 1000 & 1000 \\
AGNews  & 4 & News      & 5000 & 1000 & 1000 \\
20Newsgroup (20NG)   & 6 & News   & 3031 & 1018 & 1010 \\
Yahoo   & 6 & Q\&A       & 5000 & 1000 & 1000 \\ \bottomrule
\end{tabular}
\end{table*}

\subsubsection{Scenarios}
\label{sec:scenarios}
To see how well ICARE performs in different situations, we evaluate it in both simple and complex learning scenarios: 
\begin{itemize}[label={$\bullet$}]
    \item \textbf{Single-source domain (SSD)}: The classification model learns from data in one domain (e.g., sentiment analysis) and then adapts to a completely different target domain (e.g., Wikipedia classification). Both source and target domains have five classes.
    \item \textbf{Multi-source domains (MSD)}: The model tackles a more complex scenario with multiple sequential learning tasks. Imagine learning from sentiment analysis (Yelp + Amazon), then Wikipedia (DBPedia), then news (AGNews + 20NG), and finally question-and-answer (Yahoo). At each step, the model leverages knowledge from previous tasks while. In this scenario, at time $t$, there are $t-1$ source domains and 1 target domain. For example, at time $t = 4$, three classification tasks (e.g., sentiment, Wikipedia, and news classification) are considered as source domains, while Q\&A classification is regarded as the target domain.
\end{itemize}
\subsubsection{Model Implementation}

All experiments are executed with PyTorch library and HuggingFace Transformers frameworks \cite{wolf2020transformers} on NVIDIA RTX A5000 GPUs with 15GB memory. These models undergo training for 10 epochs with a batch size of 8 and a maximum sequence length of 256 (i.e., the first 256 tokens). The balancing hyper-parameter $\lambda$ is set to 0.5 for balancing the weights between old knowledge and new knowledge.  The learning rate is set to $1e-5$, utilizing the Adam \cite{loshchilov2017decoupled} as optimizer.
For the Class-Aware Experience Representation block, the impact of $\gamma$ ratio of exemplar samples over new data is analyzed by exploring various values (e.g., 0.01, 0.05, 0.5, 0.8). At $t = 0$, the number of old samples is insufficient for selection; thus, $\gamma$ is fixed at 0.05.
For the Knowledge Distillation block, we explore different values of the distillation temperature $\tau$, ranging from 1 to 10, to choose a suitable one. Other hyper-parameters will be presented later in appropriate sections. 

\subsubsection{Performance metrics}
\label{sec:metrics}
To assess classification model effectiveness in continual learning scenarios, we utilize several accuracy-based metrics

\begin{itemize}[label={$\bullet$}]
    \item Average Class Accuracy $ACC\textunderscore C$: This metric calculates the average accuracy across all tasks encountered during training. It provides a general sense of ICARE's overall performance.
    \item Old Class Accuracy $ACC\textunderscore Old$: This metric focuses on how well ICARE maintains accuracy for previously learned classes $C^{old}$ as it encounters new tasks. It assesses ICARE's ability to prevent catastrophic forgetting.
    \item New Class Accuracy $ACC\textunderscore New$: This metric evaluates ICARE's ability to adapt and learn new classes $C^t$ effectively in each target domain of the current task $t$.
\end{itemize}
The mathematical formula for calculating the Class Average Accuracy $ACC\textunderscore C$  over a group of categories $C$  is provided in Equation \ref{eq:avg-acc}:

\begin{equation}
    ACC\textunderscore C = Acuracy(C) = \frac{\sum_{c \in C}TP_c}{N^C}
    \label{eq:avg-acc}
\end{equation}
where $TP_c$ is the number of True Positive classification for the specific class $c$.    

It should be noted that the CDTC problem considers the overall class labels $C = C^{old} \bigcup C^t$ as the combination of old class labels  $C^{old}$ and the new classes $C^t$ learning at task $t$. Hence, the calculation for $ACC\textunderscore Old$ and  $ACC\textunderscore New$ can be done by alternatively applying Equation \ref{eq:avg-acc} for the $C^{old}$ and $C^t$ class labels. 

In addition to the previous metrics, we also employ a Forgetting metric to assess how well ICARE mitigates catastrophic forgetting. As depicted in Equation \ref{eq:forget}, this metric calculates the average difference between the best accuracy achieved on previous tasks and the average accuracy on subsequent tasks. A lower forgetting value indicates that ICARE is effectively retaining knowledge from previous tasks while learning new ones.

\begin{equation}
Forget = \frac{1}{t }\sum_{i = 1}^{t - 1} \text{max}_{i \le s < t}\{Avg(C^{old}_s)\} - Avg_{i}(C^t)
\label{eq:forget}
\end{equation}

\subsection{Results and Discussion}
\subsubsection{Evaluating Components of ICARE Frameworks }
\paragraph{Model Selection for Feature Extractor} 
For the Feature Extractor, $\mathcal{F}^t_e$, in ICARE, we explored four pre-trained language models (BERT, RoBERTa, XLNet, and DistilBERT). These models are powerful at capturing semantic information from text data.
Table \ref{tab:Backbone model} illustrates the results of ICARE with different Feature Extractors in the complex MSD scenario with a three-layer LSTM model for text classification. Overall, the RoBERTa extractor achieved the highest performance in three out of four metrics (highest $ACC\textunderscore C$: 87.58\%, highest $ACC\textunderscore New$: 80.82\%, second best $ACC\textunderscore Old$: 89.60\%, and lowest $Forget$: 8.74\%). XLNet achieves the best $ACC\textunderscore Old$ (89.71\%), indicating it might be slightly better at retaining previously learned knowledge in this complex scenario. These findings suggest that RoBERTa can offer a good balance between adapting to new data and retaining past knowledge in MSD tasks.

\begin{table*}[]
\caption{
\textbf{Impact of Feature Extractor Models.} Comparison among different state-of-the-art feature extractor models in the multi-source domains (MSD) scenario. \textbf{Boldface} indicates the best results. Values are averaged over 2 runs.}
\label{tab:Backbone model}
\centering
\begin{tabular}{@{}lcccc@{}}
\toprule
\textbf{Model} & \textbf{Avg(\%)}$\uparrow$ & \textbf{Old(\%)}$\uparrow$ & \textbf{New(\%)}$\uparrow$ & \textbf{Forget(\%)}$\downarrow$ \\ \midrule
BERT & \cellcolor[HTML]{FEF0CE}86.21 & \cellcolor[HTML]{FDECD1}86.84 & \cellcolor[HTML]{FBF2CE}80.01 & \cellcolor[HTML]{FEEBD1}10.52 \\
DistillBERT & \cellcolor[HTML]{FCE4D6}81.26 & \cellcolor[HTML]{FCE4D6}85.21 & \cellcolor[HTML]{FCE4D6}74.98 & \cellcolor[HTML]{FCE4D6}11.21 \\
RoBERTa & \cellcolor[HTML]{ECF0D5}\textbf{87.58} & \cellcolor[HTML]{E2EFDA}89.60 & \cellcolor[HTML]{E2EFDA}\textbf{80.82} & \cellcolor[HTML]{E2EFDA}\textbf{8.74} \\
XLNet & \cellcolor[HTML]{E2EFDA}86.91 & \cellcolor[HTML]{EAF0D6}\textbf{89.71} & \cellcolor[HTML]{FEF1CD}78.73 & \cellcolor[HTML]{EBEFD6}9.09 \\ \bottomrule
\end{tabular}
\end{table*}

\begin{figure}
    \centering
    \includegraphics[width=1\linewidth]{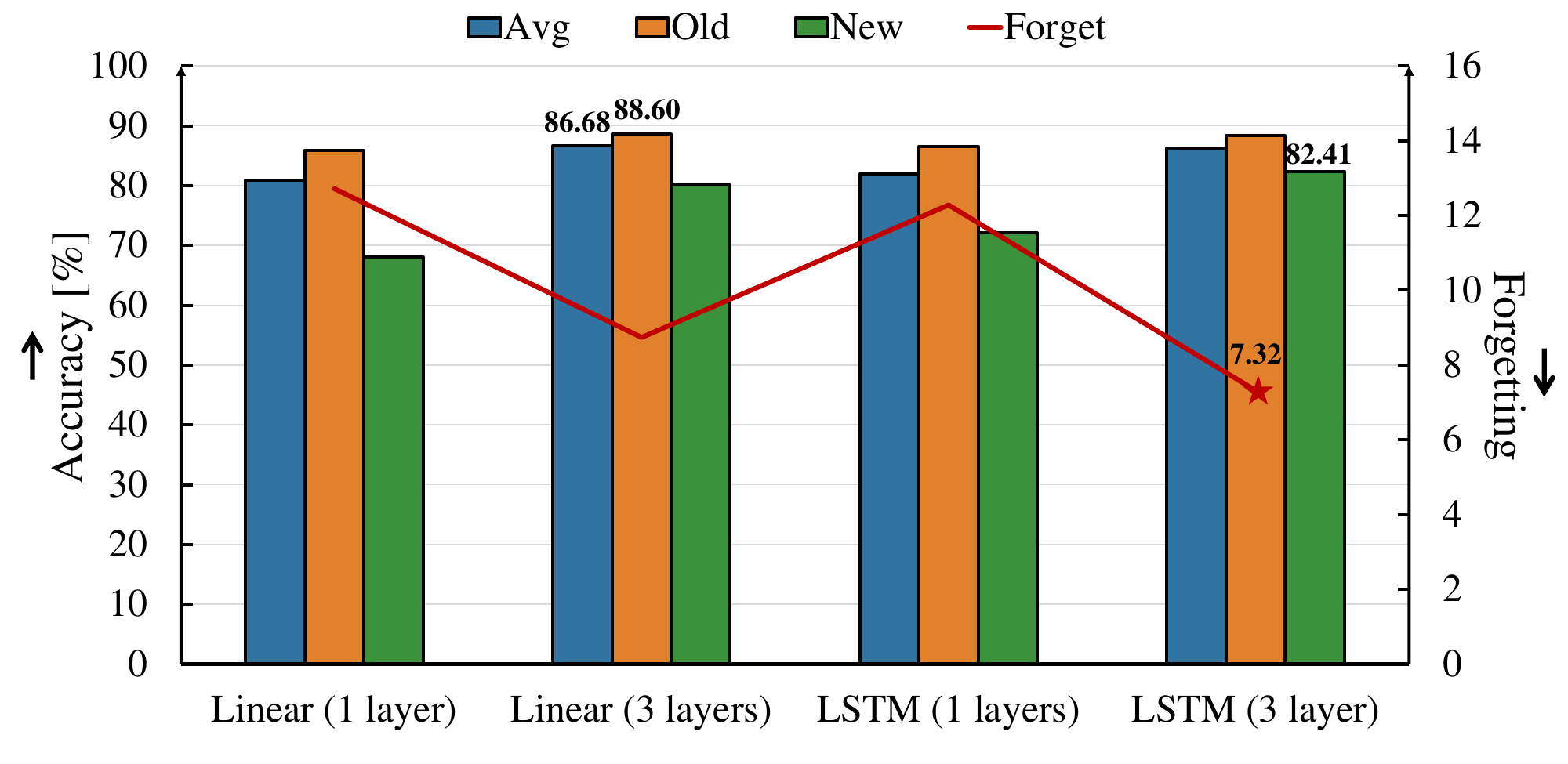}
    \caption{\textbf{Impact of Classification.} Comparison of different classifier architectures. Upward arrows ($\uparrow$) indicate higher values are preferable, whereas downward arrows ($\downarrow$) indicate lower values are preferable. Values are averaged over 2 runs.}
    \label{fig:Classification}
\end{figure}

\paragraph{Exploring classification models for the Classifier component}  
A primary component of ICARE, the classifier model $\mathcal{F}_c$, makes predictions on new data.
We explored four different models for $\mathcal{F}_c$. These models belong to two groups:
\begin{itemize}[label={$\bullet$}]
    \item Linear Models: Single-layer and three-layer Linear models have been tested. The linear models, also known as one-layer and three-layer feedforward neural networks, offer a simple and efficient approach to classification.
    \item LSTM Models: Single-layer and three-layer LSTM networks are tested. For a short recall, LSTM is a type of recurrent neural network that can effectively capture long-range dependencies within text data.
\end{itemize}

Figure \ref{fig:Classification} presents the results of testing different ICARE versions with these four classifiers. All versions use the same RoBERTa model for Feature Extractor. As per the obtained results, The three-layer linear model achieves the highest accuracy for current tasks (ACC\_C: 86.68\%) and old tasks (ACC\_Old: 88.60\%), indicating strong performance in preserving previously learned knowledge. Although it does not perform best in retaining established knowledge (ACC\_C: 86.31\%, ACC\_Old: 88.34\%), the three-layer LSTM model demonstrates superior capability in learning new classes, achieving the highest new task accuracy (ACC\_New: 82.41\%) and exhibiting the lowest forgetting rate (Forget: 7.32\%). This suggests a better balance between retaining past knowledge and adapting to new information. Considering all performance metrics, the three-layer LSTM model emerges as the preferred choice for the classifier in ICARE. It offers a good trade-off between remembering past knowledge and effectively learning new classes in continual learning scenarios.

\paragraph{Exemplar Selection Strategies: Balancing Performance and Efficiency}
This section explores the effectiveness of different exemplar selection strategies in ICARE. As described in Algorithm \ref{alg:Class-Aware Experience Representation} (refer to Section \ref{sec:Class-Aware Experience Representation} for the algorithm details), ICARE utilizes a selection process to choose representative data points (exemplars) from past tasks. Here, we compare three strategies:
\begin{itemize}[label={$\bullet$}]
    \item Random Selection: A baseline approach where exemplars are chosen randomly from the previous task's data.
    \item Rainbow Memory (Rainbow) \cite{bang2021rainbow}: This method selects exemplars based on their diversity. By applying various transformations, each exemplar is perturbed to calculate an uncertainty score, which reflects the model's response variation when processing both the original and perturbed versions. The exemplars are then ranked according to their diversity, and selected at regular intervals to ensure comprehensive coverage of the entire data space.
    \item Herding Selection (Herding): A more sophisticated approach described in Algorithm \ref{alg:Class-Aware Experience Representation} for the Class-Aware Experience Representation block. It strategically selects exemplars that can improve the model's ability to distinguish between old and new classes (refer to Algorithm \ref{alg:Class-Aware Experience Representation} for details).
\end{itemize}

Table \ref{tab:Rules} presents the results of these strategies. We evaluate them based on two key factors: the accuracy metrics and Selection Time \footnote{Selection time: the time required to select the desired exemplar samples. This measure is computed through \texttt{time.process\_time()} function in Python to capture only the actual run-time on the CPUs (i.e., sequentially).} (the selection time represents the time it takes to select the desired exemplars using each strategy.)

As the table shows, the Herding and Rainbow methods consistently outperform the Random method in terms of accuracy metrics. This suggests that both Herding and Rainbow select exemplars to achieve a better balance between retaining past knowledge and adapting to new classes. Rainbow demonstrates strong performance in $ACC\textunderscore New$, achieving a score of 81.35\%, slightly lower than the $ACC\textunderscore C$ and $ACC\textunderscore Old$ of the Herding method. Furthermore, Rainbow's Forget metric, at 8.12, indicates that this method is more effective in mitigating knowledge degradation compared to both Herding and Random.
In terms of selection time, Rainbow is significantly slower, taking approximately 450 seconds, nearly 9 times longer than Herding. This delay reflects the complexity and higher computational demands of the Rainbow method. Although Rainbow demonstrates impressive performance, especially in learning new tasks, the extended selection time makes it less suitable for time-sensitive applications compared to Herding, which offers a more efficient trade-off between accuracy and speed.

\begin{table*}[]
\caption{\textbf{Effect of Representation.} Comparison of different selection Strategies. \textbf{Boldface} indicates the best results. Values are averaged over 2 runs.}
\centering
\scalebox{0.94}{
\begin{tabular}{@{}lccccc@{}}
\toprule
\textbf{Rules} & \textbf{Avg(\%)}$\uparrow$ & \textbf{Old(\%)}$\uparrow$ & \textbf{New(\%)}$\uparrow$ & \textbf{Forget(\%)}$\downarrow$ & 
\textbf{Selection Time[s] $\downarrow$} \\ \midrule
Random         & 71.99          & 73.17          & 71.09          & 12.04           &  \textbf{$<$ 20}\\
Herding        & \textbf{87.58} & \textbf{89.60} & 80.82 & 8.74       & $\approx$ 50                                                                    \\ 
Rainbow         & 86.31          & 88.12          & \textbf{81.35}         & \textbf{8.12}     &  \textbf{$\approx$ 450}\\ \bottomrule
\end{tabular}
\label{tab:Rules}
}
\end{table*}

\begin{figure}[t]
    \centering
    \includegraphics[width=1\linewidth]{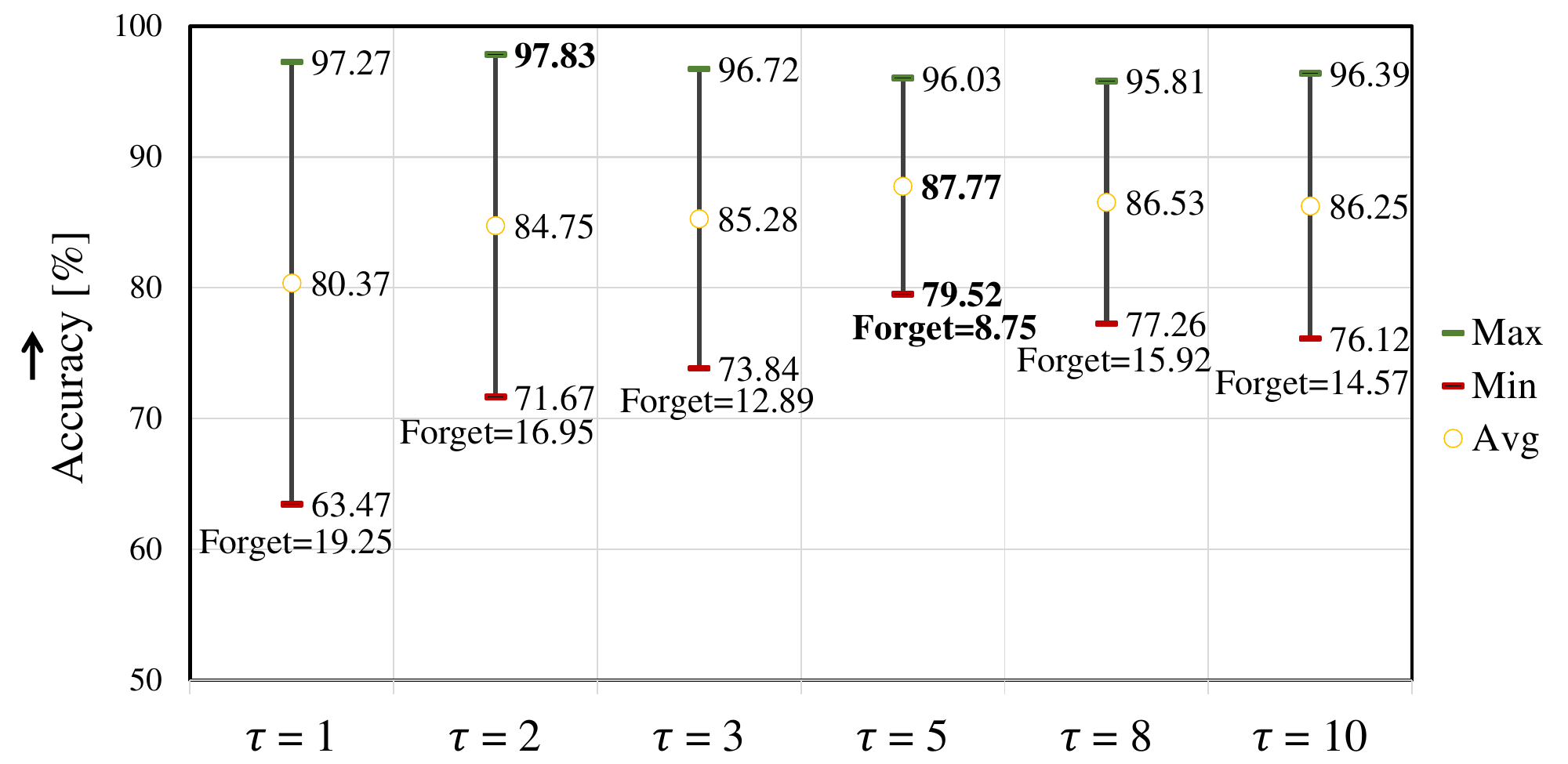}
    \caption{\textbf{Impact of Coefficients for Distillation.} Comparison of different distillation temperature values $\tau = \{1, 2, 3, 5 ,8 ,10\}$ based on the minimum, maximum, and average accuracy and forget metrics.}
    \label{fig:Distillation}
\end{figure}

\paragraph{Coefficients for Distillation.}
Knowledge distillation plays a crucial role in ICARE by transferring knowledge from the teacher model to the student model.  One key hyperparameter in this process is the distillation temperature $\tau$, which may affect the classification performance of ICARE. 

Figure \ref{fig:Distillation} illustrates how varying the distillation temperature values ($\tau = \{1, 2, 3, 5, 8, 10\}$) affects ICARE's accuracy across the accuracy and forgetting metrics.  
Based on the result, we can observe that:

\begin{itemize} [label={$\bullet$}]
    \item Balancing Forgetting and Accuracy: As the temperature parameter \(\tau\) increases from 1 to 10, both the minimum and average accuracy metrics exhibit a concave upward trend, peaking at \(\tau = 5\) with values of 79.52\% and 87.77\%, respectively. This observation suggests that an optimal temperature can effectively mitigate forgetting while preserving overall accuracy. The Forget metric further supports this finding, achieving its best value of 8.75\% at \(\tau = 5\).
    \item Max Accuracy Consistency: Interestingly, the maximum accuracy across all temperature values remains relatively stable, generally within a range of 96.03\%.
\end{itemize}

Based on these findings, we select a temperature of $\tau = 5$ as the optimal trade-off for our analyses. This value offers a good balance between mitigating forgetting and maintaining overall accuracy during knowledge distillation in ICARE.

\paragraph{Impact of Exemplar Ratio}
\begin{table*}[]
\caption{\textbf{Impact of Exemplar Ratio.} Performance metrics for varying exemplar ratio (\( \gamma \)). Values are averaged over two runs.}
\label{tab:gamma}
\centering
\scalebox{0.94}{
\begin{tabular}{@{}lccccc@{}}
\hline
\textbf{$\gamma$} & \textbf{Avg(\%)}$\uparrow$ & \textbf{Old(\%)}$\uparrow$ & \textbf{New(\%)}$\uparrow$ & \textbf{Forget(\%)}$\downarrow$ & 
\textbf{Total training time [m] $\downarrow$} \\ \hline
0.01  & \cellcolor[HTML]{FCE4D6}87.02 & \cellcolor[HTML]{FCE4D6}87.81 & \cellcolor[HTML]{E2EFDA}81.39 & \cellcolor[HTML]{FCE4D6}10.16 & \cellcolor[HTML]{E2EFDA}$\approx$ 162 \\ 
0.05  & \cellcolor[HTML]{FDE9D2}87.58 & \cellcolor[HTML]{FEEDD0}89.6  & \cellcolor[HTML]{EBF0D5}80.82 & \cellcolor[HTML]{FFF0CD}8.74  & \cellcolor[HTML]{E7F0D7}$\approx$ 168 \\
0.5   & \cellcolor[HTML]{EFF1D4}89.19 & \cellcolor[HTML]{F3F1D1}91.32 & \cellcolor[HTML]{FDEAD2}78.13 & \cellcolor[HTML]{F6F1D0}8.22  & \cellcolor[HTML]{FCE8D3}$\approx$ 235 \\
0.8   & \cellcolor[HTML]{E2EFDA}89.76 & \cellcolor[HTML]{E2EFDA}92.51 & \cellcolor[HTML]{FCE4D6}77.05 & \cellcolor[HTML]{E2EFDA}7.54  & \cellcolor[HTML]{FCE4D6}$\approx$ 281 \\ \hline
\end{tabular}
}
\end{table*}

After completing the training for the task \( t^{th} \), a subset of old samples (\( \gamma \cdot |S^t| \)) is retained for re-training during the task \( (t+1)^{th} \). Table \ref{tab:gamma} summarizes the impact of the exemplar ratio (\( \gamma = \{0.01, 0.05, 0.5, 0.8\} \)) on key performance metrics. As the exemplar ratio increases, the Forget metric decreases from 10.16 at \( \gamma = 0.01 \) to 7.54 at \( \gamma = 0.8 \), indicating the model's improved ability to retain knowledge from previous tasks. However, \( ACC\_New \) exhibits a decline, reaching its lowest value of 77.05\% at \( \gamma = 0.8 \). The average accuracy (\( ACC\_C \)) peaks at 89.76\% when \( \gamma = 0.8 \), but this comes with a substantial increase in computational cost as the Total Training Time\footnote{Total Training time: The time refers to the total duration required to execute the entire program, including the Class-Aware Experience Representation, Knowledge Distillation, and Training modules. } rises to approximately 281 minutes. These results demonstrate the influence of \( \gamma \) on the fundamental trade-off between stability (retaining existing knowledge) and plasticity (acquiring new information) in learning systems. As shown in the table, when \( \gamma = 0.05 \), we have a balance between learning new tasks and preserving the old performance ( \( ACC\_C \)) = 87.58\%, \( ACC\_Old \) =89.6\%, and \( ACC\_New \) = 80.82\%) with a low total training time (168 minutes).

Based on the obtained results, this study adopts \( \gamma = 0.05 \) as the primary configuration to evaluate the performance of the proposed ICARE model, offering a balance between experimental cost and the trade-off across key performance metrics. This observation also referred to the related works \cite{simon2022generalizing, zhao2023few, luo2023class}, where the size of the exemplar set is typically kept small, around 4 to 5\% of the training dataset to optimize memory usage and efficiency.

\subsubsection{Overall classification performance comparison}
\label{sec:Task Sequences}
\begin{figure*}[htbp]
    \centering
    \includegraphics[width=1\linewidth]{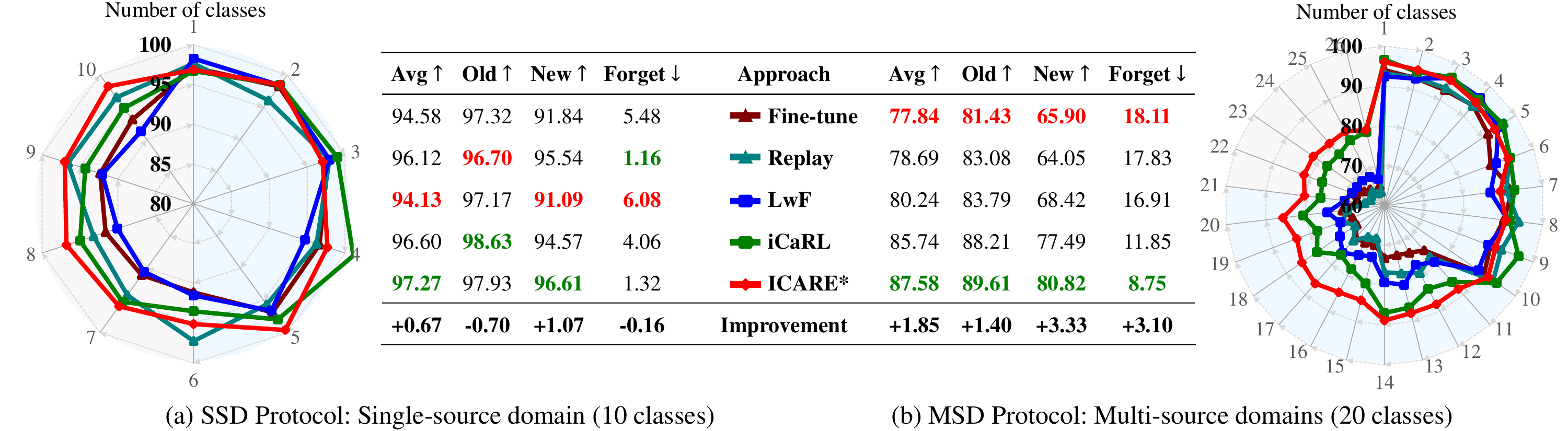}
    \caption{\textbf{Performance Evaluation.} The center table reports the per-approach metrics, with the best (\textcolor[HTML]{008000}{\textbf{bold green}}) and worst (\textcolor{red}{\textbf{bold red}}). The last row indicates the improvement of ICARE compared to the best-performing IL benchmark.}
    \label{fig:Overall}
\end{figure*}
The overall comparison of ICARE and other IL methods (Fine-tune \cite{luo2023empirical}, Replay \cite{sun2020distill}, LwF Learning without Forgetting \cite{boschini2022transfer}, and iCaRL \cite{bovenzi2021first}) on both SSD and MSD scenarios is illustrated in Figure \ref{fig:Overall}.  

For the SSD scenario, ICARE achieves the highest class average accuracy $ACC\textunderscore C$: 97.27\%) and accuracy on new classes ($ACC\textunderscore New$: 96.61\%) compared to the benchmarks. However, the Replay method performs slightly better in terms of forgetting previously learned information (Replay: 1.16\% vs. ICARE's 1.32\% ). These results suggest that ICARE demonstrates a robust capability in adapting to new classes. However, the incremental improvement in this case is relatively minor.

In the MSD scenario, ICARE demonstrates a significant advantage in this more complex scenario. The other methods seem to struggle with remembering past knowledge after training on multiple domains.  ICARE excels, particularly on new classes in the current task (target domain), achieving an accuracy of around 80.82\%, which is about 3.33\% improvement over the best-performed benchmark (iCaRL). This finding indicates that as the number of tasks increases, ICARE showcases a significant superiority over existing models. Specifically, the metrics \( ACC_{\text{New}} \) and Forget highlight ICARE's ability to effectively adapt to new classes while retaining knowledge of previously learned classes.

\subsubsection{Ablation Studies.}

We present ablation studies to investigate the effectiveness of each component in the ICARE framework. This experiment systematically removes individual components from the proposed model (ICARE) and evaluates the impact on performance. Fine-tune denotes the performance of our baseline model using the cross-entropy loss \cite{luo2023empirical}, but without the CARE, KD components, and the text classification loss $\bm{\mathcal{L}_{TC}}$, knowledge distillation loss $\bm{\mathcal{L}_{KD}}$.

\begin{table*}[htbp]
\centering
\caption{\textbf{Ablations of ICARE Components}. Comparison of the ICARE without one of the four contributions in multi-source domains (MSD) scenario. \ding{51} denotes utilization, \ding{55} denotes non-utilization. The reported values are the average of 2 runs.}
\label{tab:Componets}
\scalebox{0.82}{
\begin{threeparttable}
\begin{tabular}{@{}l|cccc|cccc@{}}
\toprule
\multicolumn{1}{c|}{\textbf{Ablation Variants}} & \textbf{CARE} & \textbf{KD} & $\bm{\mathcal{L}_{TC}}$ & $\bm{\mathcal{L}_{KD}}$ & \textbf{Avg} $\uparrow$ & \textbf{Old} $\uparrow$ & \textbf{New} $\uparrow$ & \textbf{Forget} $\downarrow$ \\ \midrule
Fine-tune & \ding{55} & \ding{55} & \ding{55} & \ding{55} & 78.01 & 83.21 & 67.51 & 17.11 \\
Fine-tune + CARE + $\mathcal{L}_{TC}$      & \ding{51} & \ding{55} & \ding{51} & \ding{51} & 80.49 & 84.12 & 71.37 & 15.21 \\
Fine-tune + KD + $\mathcal{L}_{KD}$        & \ding{55} & \ding{51} & \ding{55} & \ding{51} & 79.95 & 81.57 & 69.32 & 14.82 \\
Fine-tune + CARE + KD + $\mathcal{L}_{TC}$ & \ding{51} & \ding{51} & \ding{51} & \ding{55} & 83.72 & 87.22 & 76.91 & 9.84  \\
Fine-tune + CARE + KD + $\mathcal{L}_{KD}$ & \ding{51} & \ding{51} & \ding{55} & \ding{51} & 83.71 & 85.75 & 78.77 & 9.79  \\
ICARE (Ours)                & \ding{51} & \ding{51} & \ding{51} & \ding{51} & \textbf{87.58} & \textbf{89.60} & \textbf{80.82} & \textbf{8.75}  \\ \bottomrule
\end{tabular}

\begin{tablenotes}
\item 
\textbf{Abbreviations:} \textbf{CARE}: whether Class-Aware Experience Representation block is utilized; \textbf{KD}: whether Knowledge Distillation block is utilized; $\bm{\mathcal{L}_{TC}}$: whether Text classification loss is utilized; $\bm{\mathcal{L}_{KD}}$, whether Knowledge Distillation loss is utilized;
\end{tablenotes}
\end{threeparttable}
}
\end{table*} 

Table \ref{tab:Componets} highlights the significance of each component within the ICARE framework in the MSD scenario. Ablated models show a clear drop in performance compared to the full ICARE setup. The "Fine-tune + CARE + KD + L\_{TC}" variant achieves an $ACC\textunderscore Avg$ of 83.72\% and a Forget metric of 9.84, both of which are noticeably lower than the complete ICARE model's $ACC\textunderscore Avg$ of 87.58\% and Forget metric of 8.75. This suggests that the integration of all components (CARE, KD, \(L_{TC}\), and \(\mathcal{L}_{KD}\))is crucial for maximizing retention and adaptation in multi-domain learning.

The ICARE model (ours) demonstrates superior performance, especially with its low Forget rate of 8.75 and high $ACC\textunderscore Old$ (89.60\%) and $ACC\textunderscore New$ (80.82\%), reflects its balanced handling of stability and plasticity. This integrated approach enables ICARE to retain prior knowledge effectively while learning new classes, making it well-suited for continual learning applications in complex, evolving domains.

\section{Conclusions}
\label{sec:conclusions}
This paper proposes the Incremental Class-Aware Representation and Distillation Learning (ICARE) method for cross-domain text classification, designed to mitigate issues related to training resources and gradient vanishing/exploding. Specifically, we developed a class-aware representation module to tackle catastrophic forgetting and reduce excessive memory usage during training. We also introduced a knowledge distillation module that transfers feature-level and logit-level knowledge from the source domain to address over-parameterization issues. Our experiments highlight the superiority of ICARE over baseline methods. In the single-source domain scenario, ICARE achieved the highest class average accuracy (97.27\%) and accuracy on new classes (96.61\%) compared to four benchmarks: Fine-tune, Replay, LwF, and iCaRL. In the multi-source domain scenario, ICARE achieved an accuracy of approximately 80.82\%, surpassing the best-performing benchmark, iCaRL, by 3.33\%. Future work will focus on addressing challenges in sequence generation and sequence labeling.

\section{Acknowledgement}
The study of Son Duong was funded by the Master, PhD Scholarship Programme of Vingroup Innovation Foundation (VINIF), code VINIF.2023.ThS.113.

\bibliography{sn-bibliography}

\end{document}